\documentclass[10pt,journal,compsoc]{IEEEtran}
\ifCLASSOPTIONcompsoc
  \usepackage[nocompress]{cite}
\else
  \usepackage{cite}
\fi
\ifCLASSINFOpdf
   \usepackage[pdftex]{graphicx}
\else
\fi
\usepackage[cmex10]{amsmath}
\usepackage{graphicx}
\usepackage{caption}
\usepackage{subcaption}
\usepackage{amssymb}
\usepackage{mathabx}
\usepackage{algorithm}
\usepackage{algorithmic}
\usepackage{balance}
\newtheorem{theorem}{Theorem}[section]
\newtheorem{lemma}[theorem]{Lemma}
\newcommand\smvee{\raise0.05ex\hbox{$\scriptstyle{\vee}$}}
\newcommand\smwedge{\raise0.05ex\hbox{$\scriptstyle{\wedge}$}}

\usepackage{algorithmic}
\begin{document}
%
\title{Sparseness helps: Sparsity Augmented Collaborative Representation for Classification}
%
%
%
%

\author{Naveed~Akhtar,
        Faisal~Shafait,
        and~Ajmal~Mian
\IEEEcompsocitemizethanks{\IEEEcompsocthanksitem N. Akhtar, F. Shafait and A. Mian are with the School 
of Computer Science and Software Engineering, The University of Western Australia, 35 Stirling Highway Crawley, 6009. WA.}\\ 
\thanks{}}

%
%

\markboth{}%
{Shell}

%



\maketitle

\section*{abstract}
Many classification approaches first represent a test sample using the training samples of all the classes. This collaborative representation is then used to label the test sample. It was a common belief that sparseness of the representation is the key to success for this classification scheme. However, more recently, it has been claimed that it is the collaboration and not the sparseness that makes the scheme effective. This claim is attractive as it allows to relinquish the computationally expensive sparsity constraint over the representation. In this paper, we first extend the analysis supporting this claim and then show that sparseness explicitly contributes to improved classification, hence it should not be completely ignored for computational gains.  
Inspired by this result, we augment a dense collaborative representation with a sparse representation and propose an efficient classification method that capitalizes on the resulting representation. 
The augmented representation and the classification method work together meticulously to achieve higher accuracy and lower computational time compared to state-of-the-art collaborative representation based classification approaches. 
Experiments on benchmark face, object and action databases show the efficacy of our approach.
\begin{IEEEkeywords}
Multi-class classification, Sparse representation, Collaborative representation.
\end{IEEEkeywords}



%

\section{Introduction}
\label{sec:Int}

%
%
%
%

\IEEEPARstart{S}{everal}  
recent approaches for multi-class classification  (e.g. \cite{ESRC}, \cite{LCKSVD}, \cite{SRC}, \cite{FDDL},  \cite{CRC},  \cite{DKSVD}) exploit the representation of a test sample over a redundant basis, formed by the training samples (or their extracted features).
This \emph{collaborative representation} of the test sample, in which the training samples from different classes \emph{collaborate} to approximate the test sample, is later used to decide its  class label.
Wright et al.~\cite{SRC} first demonstrated the impressive potential of this scheme for face recognition.
Their approach additionally forces the representation to be sparse (i.e. it uses only a few vectors from the basis). Hence, it is called Sparse Representation based Classification (SRC).

The success of SRC was followed up by its variants.
For instance, Huang et al.~\cite{HH} proposed a transformation-invariant SRC.
Zhou et al.~\cite{MRFSRC} combined Markov Random Fields with SRC for disguised faces.
Similarly, Wagner et al.~\cite{PFR} enhanced SRC for the misalignment, pose and illumination invariant recognition.
Effectiveness of these approaches also boosted significant research in dictionary learning~\cite{KSVD} based multi-class classification~ \cite{DBDL}, \cite{LCKSVD}, \cite{DLCOPAR}, \cite{DLSI}, \cite{FDDL}.
Initially, the success of these approaches was  attributed to the sparseness of the used representation.
However, more recently, researchers have started questioning the role of sparsity in such approaches~\cite{AreSR}, \cite{FRLS},~\cite{CRC}.
Among them, Zhang et al.~\cite{CRC} analyzed the working mechanism of SRC and claimed that it is the \emph{collaboration and not the sparseness} of the representation that is the reason behind the effectiveness of SRC (and hence the related approaches).
This result is rather widely acclaimed as it provides grounds to relinquish the computationally expensive sparsity constraint over the representation without sacrificing the classification accuracy. 



In this paper, we first extend the analysis of Zhang et al.~\cite{CRC} and, in contrast to the original claim, we find that \emph{sparseness of collaborative representation explicitly contributes to accurate  classification}, hence it should not be completely ignored for computational gains.
Motivated by this intuition, we propose a Sparsity Augmented Collaborative Representation based Classification scheme (SA-CRC) that uses both dense and sparse collaborative representations to decide the class label of a test sample.
SA-CRC computes the dense representation using the regularized least squares method and greedily approximates the sparse representation using the Orthogonal Matching Pursuit (OMP)~\cite{OMP}. 
OMP's solution is used to augment the dense representation.
Finally, the augmented representation is classified by capitalizing on its enriched discriminative properties.  
To that end, we propose an efficient classification method that avoids explicit computation of the reconstruction residuals for each class. 
We evaluate the proposed approach on two face databases~\cite{AR}, \cite{EYaleB}, one object category database~\cite{C101} and  a dataset for action recognition~\cite{UCF}. 
Extensive experiments show that our approach is not only more accurate than the state-of-the-art collaborative representation based classification approaches, its classification time is also much lower than the approaches that ignore the sparsity altogether.

\section{Problem formulation}
\label{sec:PF}
Let $\boldsymbol\Phi \in \mathbb R^{m \times N}$ denote the training data from $C$ distinct classes, such that $ \boldsymbol\Phi = [\boldsymbol\Phi_1 ,..., \boldsymbol\Phi_i,..., \boldsymbol\Phi_C ]$. Each sub-matrix $\boldsymbol\Phi_i \in \mathbb R^{m \times n_i}$ pertains to a single class and $\sum_{i = 1}^C n_i = N$.
The columns of $\boldsymbol\Phi$ represent training samples, that are features extracted from  images.
Our goal is to develop an efficient multi-class classification scheme  by collaboratively representing a test sample ${\bf y}\in\mathbb R^m$ over the training data\footnote{No explicit training of a machine learning algorithm is aimed, $\boldsymbol\Phi$ is conventionally  referred as the \emph{training} data~\cite{SRC}, \cite{CRC}.}. 
A test sample is considered to be a feature vector that can be  linearly  approximated by the training samples.
That is, ${\bf y} \approx \boldsymbol\Phi \boldsymbol\alpha$, where $\boldsymbol\alpha\in \mathbb R^{N}$ is the Collaborative Representation~(CR) vector of the test sample.
We allow $\boldsymbol\Phi$ to be a redundant set of basis vectors   in $\mathbb R^m$.
Furthermore, the subspaces spanned by the sub-matrices $\boldsymbol\Phi_{i \in \{1,...,C\}}$ are  considered to be possibly overlapping, as this is often the case for the multi-class classification problems.
Following the sparse representation literature~\cite{Proc}, \cite{Proc1}, we alternatively refer to  $\boldsymbol\Phi$ as the  \emph{dictionary} and to its columns as the \emph{dictionary atoms}. 
Furthermore, we generally refer to the \emph{representation vector} (e.g. $\boldsymbol\alpha$) as \emph{representation}, for brevity. 

\section{Related work}
\label{sec:RW}
 \floatname{algorithm}{Algorithm}
\renewcommand{\algorithmicrequire}{\textbf{Initializaiton:}}
\renewcommand{\algorithmicensure}{\textbf{Main Iteration:}}
\floatname{algorithm}{Algorithm}
\renewcommand{\algorithmicrequire}{\textbf {Input:}}
\renewcommand{\algorithmicensure}{\textbf{Output:}}
\begin{algorithm}
\caption{CR-based Classification} 
\begin{algorithmic}[1]
\REQUIRE (a) Training data $\boldsymbol\Phi$, with samples  normalized to have unit $\ell_2$-norm. (b) Test sample ${\bf y}$. (c)  Regularization parameter $\lambda$.
\STATE  \emph{Optimization:} Solve 
\begin{align}
\boldsymbol\alpha = \min_{\alpha} || {\bf y} - \boldsymbol\Phi \boldsymbol\alpha ||_2^2 + \lambda f(\boldsymbol\alpha),
\label{eq:opt1}
\end{align}
where, $f(.)$ denotes a function and $||.||_p$ represents the $\ell_p$-norm of a vector. 
\STATE \emph{Residual computation:}  Compute class-specific reconstruction residuals $r_i({\bf y}) = || {\bf y} - \boldsymbol\Phi_i \acute{\boldsymbol\alpha_i}||_2,~\forall i\in\{1,...,C\}$, where $\acute{\boldsymbol\alpha_i}\in \mathbb R^{n_i} $ comprises the coefficients of $\boldsymbol\alpha$ corresponding to the $i^{\text{th}}$ class. 
\STATE \emph{Labeling:} label$({\bf y}) = \min_{i} \{r_i({\bf y})\}$. 
\ENSURE  label$({\bf y})$. 
\end{algorithmic} 
\label{alg:1}
\end{algorithm}

Algorithm~\ref{alg:1} presents the base-line scheme used by popular approaches (e.g. \cite{SRC}, \cite{SSRC}, \cite{CRC}, \cite{FRLS}, \cite{DLCOPAR}, \cite{FDDL}) that exploit     collaborative representation in multi-class classification.
The algorithm performs three key steps of (1)~optimizing ${\bf y}$'s   representation over a given dictionary, (2)~computing class-specific  reconstruction residuals $r_i({\bf y})$, $\forall i \in \{1,...,C\}$ and  (3)~labeling ${\bf y}$ using the computed residuals. 
In step~(2),  $\acute{\boldsymbol\alpha_i} \in \mathbb R^{n_i}$ comprises the coefficients of $\boldsymbol\alpha$ corresponding to the $i^{\text{th}}$ class only.
Hence, in step~(3), ${\bf y}$ is assigned the label of the class that results in the smallest reconstruction residual. 
We can treat different approaches as special cases of the presented  algorithm. 

In SRC~\cite{SRC}, $f(\boldsymbol\alpha) = ||\boldsymbol\alpha ||_1$ in Eq.~(\ref{eq:opt1}), which encourages the computed representation $\boldsymbol\alpha$ to be sparse. 
In Superposed-SRC (SSRC), Deng et al.~\cite{SSRC} modified the  residual computation step of SRC. 
For SSRC, $\boldsymbol\Phi$ consists of class centroids and sample-to-centroid differences.
While computing the residuals, SSRC keeps the coefficients of $\boldsymbol\alpha$ corresponding to the sample-to-centroid differences fixed in each $\acute{\boldsymbol\alpha_i}$.   
The CR-based classifier proposed by Zhang et al.~\cite{CRC} uses  $f(\boldsymbol\alpha) = ||\boldsymbol\alpha||_2$ and solves Eq.~(\ref{eq:opt1}) using the Regularized Least Squares (RLS) method, hence denoted as CRC-RLS. 
Shi et al.~\cite{FRLS} used $\lambda = 0$ in Eq.~(\ref{eq:opt1}) and solved it as the standard least squares problem for face recognition.   
Chi and Porikli~\cite{CROC} used a linear combination of a CR-based classifier and a nearest subspace classifier~\cite{NSC} for improved classification performance.

Collaborative representation is also commonly used by   discriminative dictionary learning techniques, e.g.~\cite{DLCOPAR}, \cite{FDDL}. 
Although such approaches \emph{learn} a dictionary instead of  directly using the training data as $\boldsymbol\Phi$, explicit  correspondence between the learned dictionary atoms and the class labels allows them to exploit the CR-based classification scheme. 
For instance, the Global Classifier (GC) used by Kong and Wang~\cite{DLCOPAR} is the same variant of Algorithm~\ref{alg:1} that is used by SSRC~\cite{SSRC}. 
The dictionary learned by the DL-COPAR algorithm \cite{DLCOPAR} consists of COmmon atoms for all classes and PARticular atoms specific to each class. 
The particular atoms behave like class centroids whereas the common atoms act as centroid-to-sample differences in SSRC.  
Similarly, the GC used in the Fisher Discriminant Dictionary Learning (FDDL)~\cite{FDDL} is a direct variant of CRC-RLS~\cite{CRC}.


Another interesting direction of discriminative dictionary learning techniques, e.g. Label Consistent K-SVD (LC-KSVD)~\cite{LCKSVD} and Discriminative K-SVD (D-KSVD)~\cite{DKSVD},  is also related to CR-based classification.
Such techniques learn collaborative dictionaries from the training data without enforcing strict correspondence between the class labels and the dictionary atoms.
Due to the lack of such correspondence, the label of a test sample is chosen by maximizing a weighted  sum of the coefficients of $\boldsymbol\alpha$, where
the $N$-dimensional $C$ weight-vectors are also learned during  dictionary optimization. 
Among these weight-vectors, the $i^{\text{th}}$ vector generally assigns large weights to the coefficients of $\boldsymbol\alpha$ corresponding to the dictionary atoms used commonly in representing the training data of the $i^{\text{th}}$ class. 

The above mentioned discriminative dictionary learning approaches classify a test sample using its representation over a collaborative set of features, learned directly from the training data. 
Therefore, they are considered to be instances of CR-based classification.

\section{Collaboration and Sparsity}
\label{sec:CS}
It is clear from Section~\ref{sec:RW} that many popular approaches directly exploit collaborative representation $\boldsymbol\alpha$  in classification. 
Whereas sparse representation based  approaches (e.g. \cite{SRC}, \cite{SSRC})  associate the discriminative power of $\boldsymbol\alpha$ to its sparseness, there is an equal evidence in favor of discriminative abilities of dense representations  \cite{CROC},  \cite{FRLS}, \cite{CRC}. 
In fact, it is also advocated that sparsity of the representation may not even be relevant to classification \cite{AreSR}, \cite{FRLS}, \cite{CRC}. 
Zhang et al.~\cite{CRC} boosted the popularity of this notion by corroborating their claim with an analysis of the working mechanism of SRC. 
In Section~\ref{sec:CR}, we closely follow this analysis to explain the role of collaboration in CR-based classification.
We extend this analysis on the same lines of reasoning in Section~\ref{sec:CRfails} to show that collaboration alone is not sufficient for accurate classification.
Section~\ref{sec:SR} discusses how sparseness additionally helps in this regard. 


\subsection{Why collaboration works?}
\label{sec:CR}
\begin{figure}[t]
        \centering
        \begin{subfigure}[b]{0.35\textwidth}
                \includegraphics[width=\textwidth, height = 1.6in]{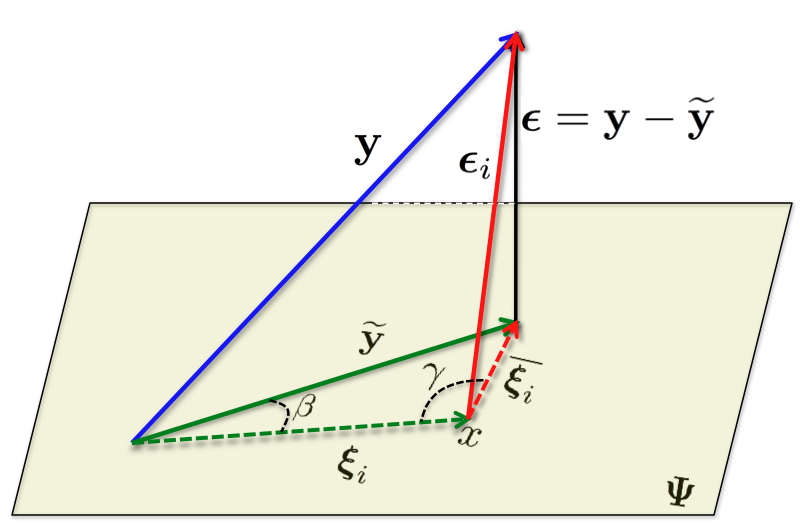}
                \caption{}
                \label{fig:ill1a}
	\end{subfigure}\\
\centering
        \begin{subfigure}[b]{0.35\textwidth}
                \includegraphics[width=\textwidth, height = 1.6in]{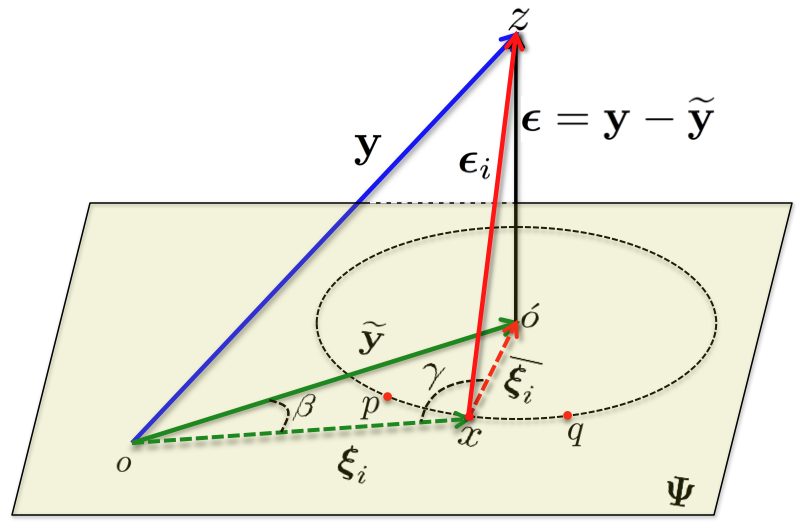}
                \caption{}
                \label{fig:ill1b}
	\end{subfigure}
   \caption{Geometric illustration of the working mechanism of CR-based classification.}
   \label{fig:ill1}
\end{figure}

We write the subspace spanned by the columns of $\boldsymbol\Phi$ as a set $\boldsymbol\Psi$. 
This subspace is geometrically illustrated as a plane in Fig.~\ref{fig:ill1}. 
Since a test sample ${\bf y}$ is approximated by the columns of $\boldsymbol\Phi$, we can write the approximation error as $\boldsymbol\epsilon = {\bf y} - \widetilde{\bf y}$, where $\widetilde{\bf y} = \boldsymbol\Phi \boldsymbol\alpha \subset \boldsymbol\Psi$
\footnote{For $\boldsymbol\Phi \in \mathbb R^{m\times N}$, ${\bf y} \subset \boldsymbol\Psi$ when $N\rightarrow \infty$ and $\boldsymbol\epsilon \perp \widetilde{\boldsymbol\Psi}  $, where $\widetilde{\boldsymbol\Psi}~\subset~{\boldsymbol\Psi}$.
In that case, we are concerned with $\widetilde{\boldsymbol\Psi}$ only, as ${\bf y}$ is considered to be approximated with a small error of bounded energy, i.e. $||\boldsymbol\epsilon||_2~\leq~\varepsilon$.
We exaggerate the error vector in figures for clarity.}.  
Let us represent the subspace spanned by the training data of the $i^{\text{th}}$ class by a set $\boldsymbol\Psi_i$, where $\boldsymbol\Psi_i \subset \boldsymbol\Psi$. 
Without loss of generality, we can decompose $\widetilde{\bf y}$ into two components, $\boldsymbol\xi_i$ and $\overline{\boldsymbol\xi_i}$ (illustrated in Fig.~\ref{fig:ill1a}) such that $\boldsymbol\xi_i \subset \boldsymbol\Psi_i$ and $\overline{\boldsymbol\xi_i} \subset \overline{\boldsymbol\Psi_i}$,  where $\overline{\boldsymbol\Psi_i} = \bigcup_{j = 1; j\neq i}^C \boldsymbol\Psi_j$.
Similarly, the total approximation error $\boldsymbol\epsilon$ can itself be considered as a component of $\boldsymbol\epsilon_i$, where $||\boldsymbol\epsilon_i ||_2$ represents the class-specific reconstruction  residual $r_i({\bf y})$, see step~$2$ of Algorithm~\ref{alg:1}. 

To understand the working mechanism of CR-based classification, let ${\bf y}$ belong to the $c^{\text{th}}$ class. 
In this case,  $\widetilde{\bf y} = \boldsymbol\xi_c + \overline{\boldsymbol\xi_c}$, i.e. $i = c$ in Fig.~\ref{fig:ill1a}.
A CR-based classifier selects $c$ as the label of ${\bf y}$ because $\boldsymbol\epsilon_{i}$ is expected to have the smallest length when $i=c$~\cite{SRC}, \cite{CRC}. 
Zhang et al.~\cite{CRC} noted that this labeling criterion not only considers that the angle between $\widetilde{\bf y}$ and $\boldsymbol\xi_c$ (i.e. $\beta$) is small, it also considers that the angle between $\boldsymbol\xi_c$ and $\overline{\boldsymbol\xi_c}$ (i.e. $\gamma$) is large. 
According to Zhang et al.~\cite{CRC}, it is this \emph{double-check with $\beta$ and $\gamma$} (not the sparseness of the representation) that makes CR-based classification robust and effective.
Therefore, they solved Eq.~(\ref{eq:opt1}) using a computationally efficient regularized least squares method. 
The resulting dense collaborative representation was shown to be effective for face recognition, similar to sparse representation. 



\subsection{Why collaboration alone is not sufficient?}
\label{sec:CRfails}
In the following text, we refer to a vector $\boldsymbol\epsilon_i$ as \emph{class-specific error vector}.
We present Lemma~\ref{lem:1} regarding the underlying geometry of the class-specific error vectors involved in CR-based classification:

\begin{lemma}
\label{lem:1}

For $i,j,k \in \{1,...,C\}$, where $i \neq j \neq k$, the following holds: $\exists~\boldsymbol\epsilon_i, \boldsymbol\epsilon_j$ such that $||\boldsymbol\epsilon_i||_2 = ||\boldsymbol\epsilon_j||_2$, while $\nexists~\boldsymbol\epsilon_k$ such that $||\boldsymbol\epsilon_k||_2 < ||\boldsymbol\epsilon_i||_2$.
\end{lemma}
{\bf Proof:}
For our problem, the following holds under the law of sines, which can be verified from Fig.~\ref{fig:ill1a}:
\begin{align}
\frac{||\widetilde{\bf y}||_2}{\sin(\gamma)} = \frac{||\overline{\boldsymbol\xi_i}||_2}{\sin(\beta)}.
\label{eq:2}
\end{align}
Also, $|| \boldsymbol\epsilon_i||_2^2 = ||\boldsymbol\epsilon||_2^2 + ||\overline{\boldsymbol\xi_i}||_2^2$ because $\boldsymbol\Psi \perp  \boldsymbol\epsilon$. 
From Eq.~(\ref{eq:2}),
\begin{align}
||\boldsymbol\epsilon_i||_2^2 = ||\boldsymbol\epsilon||_2^2 + \left(\frac{\sin(\beta)}{\sin(\gamma)}\right)^2 ||\widetilde{\bf y}||_2^2.
\label{eq:3}
\end{align}
Since $||\boldsymbol\epsilon||_2^2$ and $||\widetilde{\bf y}||_2^2$ become constants once ${\bf y}$ is projected onto $\boldsymbol\Psi$, the condition that $\nexists~\boldsymbol\epsilon_k$ s.t. $ ||\boldsymbol\epsilon_k||_2 < ||\boldsymbol\epsilon_i||_2$ holds when $\big(\sin (\beta) / \sin(\gamma)\big)^2$ is the minimum.
However, for $\beta,\gamma \in [0, 2\pi]$ there is no unique minima for the given squared ratio. 
Hence, it is possible that $\exists~\boldsymbol\epsilon_i, \boldsymbol\epsilon_j$ s.t. $||\boldsymbol\epsilon_i||_2 = ||\boldsymbol\epsilon_j||_2$, while $\nexists~\boldsymbol\epsilon_k$ s.t. $||\boldsymbol\epsilon_k||_2 < ||\boldsymbol\epsilon_i||_2$.

Lemma~\ref{lem:1}, shows the possibility of existence of multiple class-specific error vectors with equal lengths when the length is minimized over the class labels. 
Figure~\ref{fig:ill1b} illustrates this possibility  by drawing a circle of radius $||\overline{\boldsymbol\xi_i}||_2$ around point $\acute{o}$ on $\boldsymbol\Psi$. 
Any vector starting from a point on this circle (e.g. $p,q$) and ending at $z$ will have the same length. 
For the labeling criterion of CR-based classification scheme, collaboration of the representation alone is  not sufficient to indicate the best vector among these possible vectors.  
From Lemma~\ref{lem:1}, it is also evident that the double-check with $\beta$ and $\gamma$ mentioned by Zhang et al.~\cite{CRC} is essentially a \emph{single-check} on the squared ratio of the sines of the angles.
Thus, \emph{CR-based classification without considering sparsity  may not be as robust and effective as previously thought}.



\subsection{How sparseness helps?}
\label{sec:SR}
The above mentioned issue is inherent to CR-based classification scheme, with its roots in the redundancy in  ${\boldsymbol\Phi}$. 
Simply computing a unique approximation of the representation, such as in CRC-RLS~\cite{CRC}, does not resolve the issue because Lemma~\ref{lem:1} still holds for the labeling step in Algorithm~\ref{alg:1}. 
To truly address the problem, a collaborative representation must be infused with additional information that finally results in using a suitable class-specific error vector in the labeling step. 
Sparsity constraint over the representation serves this purpose in CR-based classification.



\begin{figure}[t]
        \centering
        \begin{subfigure}[b]{0.35\textwidth}
                \includegraphics[width=\textwidth, height = 1.6in]{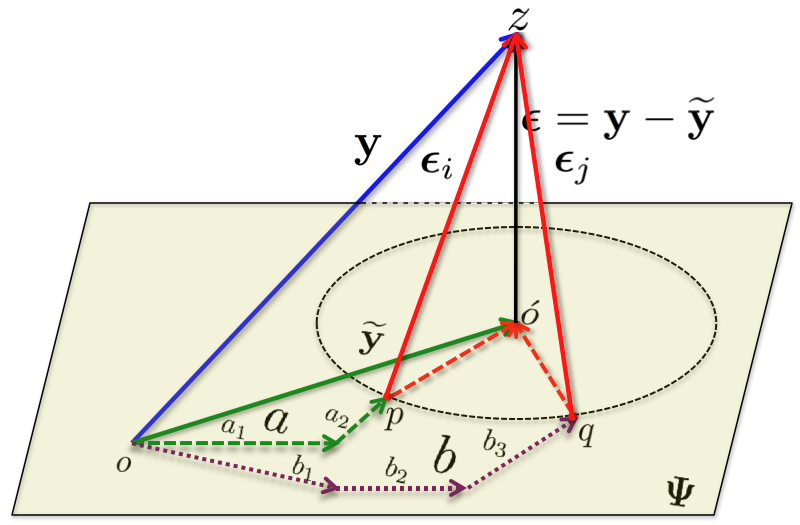}
                \caption{}
                \label{fig:ill2a}
        \end{subfigure}\\%
        \begin{subfigure}[b]{0.35\textwidth}
                \includegraphics[width=\textwidth, height = 1.6in]{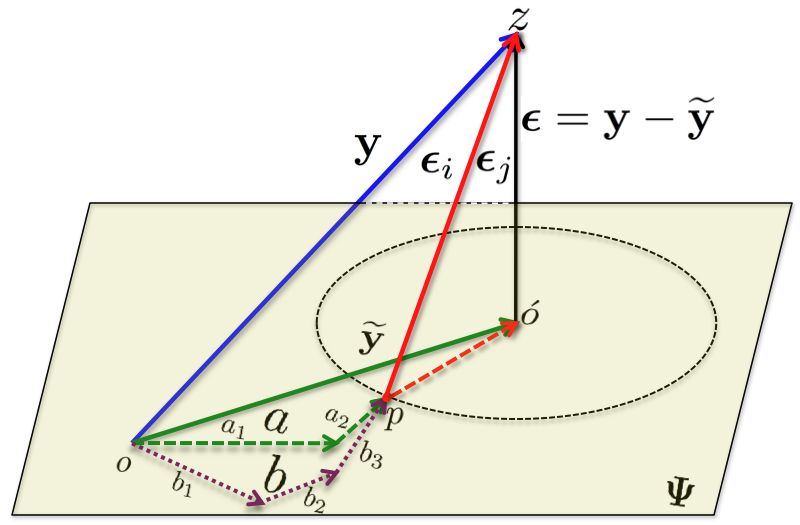}
                \caption{}
                \label{fig:ill2b}
        \end{subfigure}
        \caption{Geometric illustration of the jointly exhaustive cases when $\exists \boldsymbol\epsilon_i, \boldsymbol\epsilon_j $ such that $||\boldsymbol\epsilon_i||_2 = ||\boldsymbol\epsilon_j||_2$: (a)~$\boldsymbol\xi_i \neq \boldsymbol\xi_j$. (b)~$\boldsymbol\xi_i = \boldsymbol\xi_j$. Here, $\boldsymbol\xi_i = {\bf a} $ and $\boldsymbol\xi_j = {\bf b}$ and the vectors are only displayed in terms of their components.}
        \label{fig:ill2}
\end{figure}

To support our argument, in Fig.~\ref{fig:ill2}, we geometrically illustrate the two jointly exhaustive situations that can occur when two class-specific error vectors $\boldsymbol\epsilon_i$ and $\boldsymbol\epsilon_j$ have equal lengths, namely (a)~$\boldsymbol\xi_i \neq \boldsymbol\xi_j$ and  (b)~$\boldsymbol\xi_i = \boldsymbol\xi_j$.
In the figure, we denote $\boldsymbol\xi_i$ by ${\bf a}$ 
and $\boldsymbol\xi_j$ by ${\bf b}$ 
and show these vectors only by their components to avoid cluttering.
In Fig.~\ref{fig:ill2a}, ${\bf a} \neq {\bf b}$ but $||\boldsymbol\epsilon_i||_2 = ||\boldsymbol\epsilon_j||_2$. 
In Fig.~\ref{fig:ill2b}, ${\bf a} = {\bf b} = \overrightarrow{op}$ and $||\boldsymbol\epsilon_i||_2 = ||\boldsymbol\epsilon_j||_2$. 
Although the class-specific residuals are equal in both cases, $\boldsymbol\xi_i$ and $\boldsymbol\xi_j$ can be distinguished based on their components.
Intuitively, $i$ (not $j$) represents the correct class of the test sample because $\boldsymbol\xi_i$ requires lesser number of components to produce the smallest class-specific residual.
Fewer components of $\boldsymbol\xi_i$   
implicates a sparser $\boldsymbol\alpha$.
Hence, the sparsity constraint  results in using a better class-specific error vector in the labeling step. 
Incidentally, the best performance of CR-based classification can be achieved by guaranteeing the representation to be the sparsest possible.

\section{Proposed approach}
\label{sec:PA}

Computing the sparsest possible representation is generally NP-hard~\cite{NPHard}. 
SRC~\cite{SRC} uses the $\ell_1$-norm constraint to compute an approximate sparse representation, but the approach remains computationally expensive. 
On the other hand, computing a dense representation, such as  in CRC-RLS~\cite{CRC}, resolves the computational issues but it does not offer the advantages of sparsity.
In the proposed classification scheme, we augment a dense representation with a greedily obtained approximate sparse representation.
This augmentation enables accurate classification 
while keeping the approach computationally efficient.

Algorithm~\ref{alg:2} presents the proposed scheme.
In the first step, the algorithm optimizes two collaborative representations, i.e. $\boldsymbol{\widecheck\alpha}$ and $\widehat{\boldsymbol\alpha}$.
The dense representation $\boldsymbol{\widecheck\alpha}$ is computed using the regularized least squares method, whereas the sparse representation $\widehat{\boldsymbol\alpha}$ is obtained by solving Eq.~(\ref{eq:opt2}) using the Orthogonal Matching Pursuit (OMP) algorithm~\cite{OMP}.
OMP iteratively selects $k$ dictionary atoms to represent ${\bf y}$, hence,  $\widehat{\boldsymbol\alpha}$ has at most $k$ non-zero coefficients, where $k$ (the sparsity threshold)  is determined by cross-validation.  
In each iteration, OMP chooses a dictionary atom by maximizing its  correlation with an error vector. 
The error vector is computed as the difference between ${\bf y}$ and its orthogonal projection onto the subspace spanned by the already chosen atoms. 
For initialization, ${\bf y}$ itself is considered as the error.

As shown in step~$2$ of Algorithm~\ref{alg:2}, we add the sparse representation $\widehat{\boldsymbol\alpha}$ to $\boldsymbol{\widecheck\alpha}$ and normalize the resulting vector to compute the augmented representation~$\boldsymbol{\overset{\circ}\alpha}$.
Despite being simple, this procedure greatly improves the discriminative abilities of the representation. 
We defer the discussion on the discriminative properties of $\boldsymbol{\overset{\circ}\alpha}$ to the upcoming paragraphs.
These properties are exploited in step~(3) of the algorithm to efficiently compute the label of the test sample~${\bf y}$. 
The labeling step uses a binary matrix ${\bf L}\in \mathbb R^{C \times N}$, that is provided as an input to the algorithm.
For the $i^{\text{th}}$ class, ${\bf L}$ contains $n_i$ non-zero elements in its $i^{\text{th}}$ row, at the indices corresponding to the columns of $\boldsymbol\Phi_i$.
Thus, the $i^{\text{th}}$ coefficient of $ {\bf q} = {\bf L}\overset{\circ}{\boldsymbol\alpha}$ represents the sum of $\overset{\circ}{\boldsymbol\alpha}$'s coefficients corresponding to  $\boldsymbol\Phi_i$. 
The label of the test sample is decided by maximizing the coefficients of ${\bf q}$. 
Empirical evidence for efficient and accurate classification using the proposed scheme is provided  
in Sections~\ref{sec:Exp}. 
Below, we analyze the reasons behind the improved performance of the approach. 

\floatname{algorithm}{Algorithm}
\renewcommand{\algorithmicrequire}{\textbf{Initializaiton:}}
\renewcommand{\algorithmicensure}{\textbf{Main Iteration:}}
\floatname{algorithm}{Algorithm}
\renewcommand{\algorithmicrequire}{\textbf {Input:}}
\renewcommand{\algorithmicensure}{\textbf{Output:}}
\begin{algorithm}
\caption{Sparsity Augmented CR-based Classification} 
\begin{algorithmic}[1]
\REQUIRE (a) Training data $\boldsymbol\Phi$, with samples  normalized in $\ell_2$-norm. (b) Test sample ${\bf y}$. (c) Regularization parameter~$\lambda$. (d)~Sparsity threshold $k$. (e) Label matrix ${\bf L}$.
\\
\STATE  \emph{Optimization:}\\
a)~Compute $
\boldsymbol{\widecheck\alpha} = {\bf P}{\bf y},
\label{eq:RLS}
$
where, ${\bf P} = (\boldsymbol\Phi^{\text{T}} \boldsymbol\Phi + \lambda{\bf I}_N)^{-1}\boldsymbol\Phi^{\text{T}}$.
b)~Solve the following with greedy pursuit:
\begin{align}
\widehat{\boldsymbol\alpha} = \min_{\alpha} ||{\bf y} - \boldsymbol\Phi \boldsymbol\alpha||_2,~s.t.~||\boldsymbol\alpha||_0 \leq k,
\label{eq:opt2}
\end{align}
where, $||.||_0$ denotes the $\ell_0$-pseudo norm.
\STATE \emph{Augmentation:}  Compute 
\begin{align}
{\boldsymbol{\overset{\circ}\alpha}} = \frac{\widehat{\boldsymbol\alpha} + \boldsymbol{\widecheck\alpha}}{||\widehat{\boldsymbol\alpha} + \boldsymbol{\widecheck\alpha}||_2}
\label{eq:alpha}
\end{align}
\STATE \emph{Labeling:} label$({\bf y}) = \text{arg max}_{i}\{q_i\}$, where $q_i$ denotes the $i^{\text{th}}$ coefficient of ${\bf q} = {\bf L}\boldsymbol{\overset{\circ}\alpha}$. 
\ENSURE  label$({\bf y})$. 
\end{algorithmic} 
\label{alg:2}
\end{algorithm}

 \begin{figure*}[t] 
    \centering
    \includegraphics[width=7in, height=2.3in]{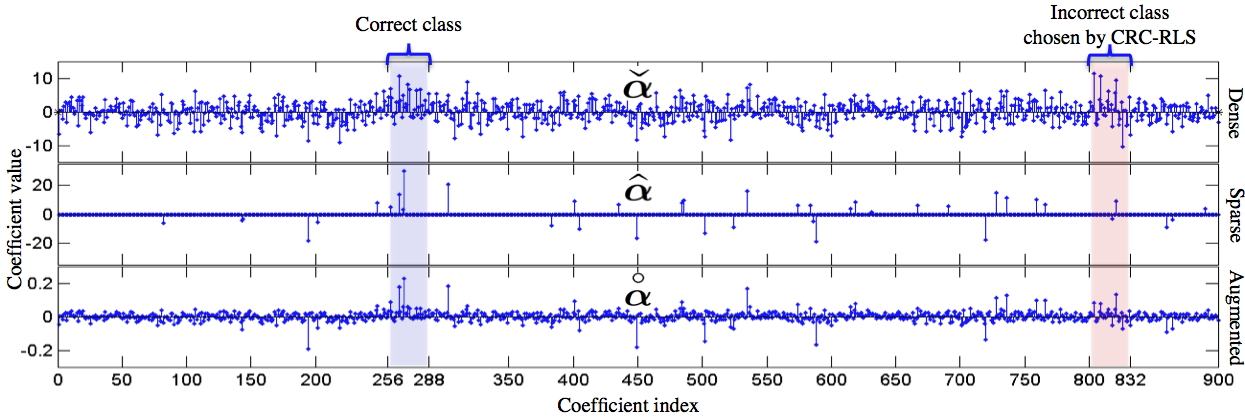} 
    \caption{Comparison of a test sample's representations for face recognition task using Extended YaleB database~\cite{EYaleB}: $\widehat{\boldsymbol\alpha}$ consistently shows large positive values at the coefficients corresponding to the correct class. This culminates in correct classification by SA-CRC, whereas CRC-RLS~\cite{CRC} chooses incorrect class label for the test sample despite optimized parameters. Only first 900 coefficients are shown out of 1216.}
    \label{fig:SC}
 \end{figure*}

For analysis, let us distribute the coefficient indices of a    collaborative representation $\boldsymbol\alpha$ into two disjoint sets: $\mathcal{A_H} = \{ i : 
\Xi_i > \delta\}$
and  $\mathcal{A_L} = \{j : 
\Xi_j \leq \delta \}$, where $\delta \in \mathbb R^+$ and $\Xi_n = \frac{\alpha_n^2}{||\boldsymbol\alpha||_2^2}$ with $\alpha_n \in \mathbb R$ denoting the $n^{\text{th}}$ coefficient of $\boldsymbol\alpha$.
The value $\Xi_n$ represents the energy 
 in the $n^{\text{th}}$ coefficient, 
such that  $\sum\limits_{n=1}^N \Xi_n = 1$. 
If $\delta = 0$,  $\mathcal{A_H}$ contains the indices of non-zero coefficients of $\boldsymbol\alpha$, whereas  $\mathcal{A_L}$ comprises the indices of zero coefficients. 
Thus, the cardinality of the set $\mathcal{A_H}$, i.e. $|\mathcal{A_H}|$, defines the sparsity level of~$\boldsymbol\alpha$.
This remains true for $0 \leq \delta < \min\limits_i~\Xi_i$.
Let $\boldsymbol\alpha^*$ denote the sparsest possible representation of ${\bf y}$ over $\boldsymbol\Phi$.  
We write the aforementioned sets for $\boldsymbol\alpha^*$ as $\mathcal{A_H^*}$ and $\mathcal{A_L^*}$.
Furthermore, for any $\boldsymbol\alpha$, 
let us now fix $\delta =  \left(\frac{\alpha^*_{min}}{||\boldsymbol\alpha^*||_2}\right)^2 - \varepsilon$, where $\alpha^*_{min}$ denotes the lowest energy coefficient of $\boldsymbol\alpha^*$.
Hence, $|\mathcal{A_H}|$ now counts the number of coefficients of $\boldsymbol\alpha$, each having at least the energy possessed by $\alpha^*_{min}$.
Therefore, henceforth, we refer to $|\mathcal{A_H}|$ as the \emph{effective sparsity} of the representation. 






From Section~\ref{sec:CS}, we know that $\boldsymbol\alpha^*$ is  discriminative due to its sparsity.   
In practice, a representation $\overset{\circ}{\boldsymbol\alpha}$ is  equally effective for classification if $|\mathcal{A_H^{\circ}}|\approx |\mathcal{A^*_H}|$ and the coefficients indexed in 
 $\mathcal{A_H^{\circ}}$ are discriminative\footnote{We can safely ignore $\mathcal{A_L^{\circ}}$ in this argument because the coefficients indexed in $\mathcal{A_L^{\circ}}$ can be explicitly forced to zero, once $\mathcal{A_H^{\circ}}$ is known.}. 
For a dense representation $\boldsymbol{\widecheck\alpha}$, $|\mathcal{A_H^{\smvee}}| \approx N \gg |\mathcal{A_H^*}|$. 
Nevertheless, the representation is globally optimal. 
On the other hand, $|\mathcal{A_H^{\smwedge}}| \approx k \ll N$ for the sparse representation $\widehat{\boldsymbol\alpha}$, but the representation is only locally optimal.
However, $\widehat{\boldsymbol\alpha}$ generally contains large  positive coefficients at the indices corresponding to the correct class.
For the other classes, most of the coefficients are either negative or have small positive values. 
This happens because OMP greedily assigns large values to the coefficients of $\widehat{\boldsymbol\alpha}$ corresponding to the dictionary atoms that correlate more to ${\bf y}$, whereas ${\bf y}$ generally has a strong positive correlation with the samples of its own class.
Thus, adding $\widehat{\boldsymbol\alpha}$ to $\boldsymbol{\widecheck\alpha}$ amplifies the coefficients of the correct class in the globally optimal solution. 
Figure~\ref{fig:SC} illustrate this phenomenon using an actual example of face recognition. 
In the figure, the coefficients of $\widehat{\boldsymbol\alpha}$ are consistently positive and have relatively large values for the correct class. 
This finally results in dominant positive coefficients of $\overset{\circ}{\boldsymbol\alpha}$ for the correct class.  
For this example, CRC-RLS~\cite{CRC} is not able to identify the correct label of ${\bf y}$ despite optimized parameter settings, whereas the proposed approach classifies ${\bf y}$ correctly.

\begin{figure}[t] 
    \centering
    \includegraphics[width=2.4in, height = 1.5in]{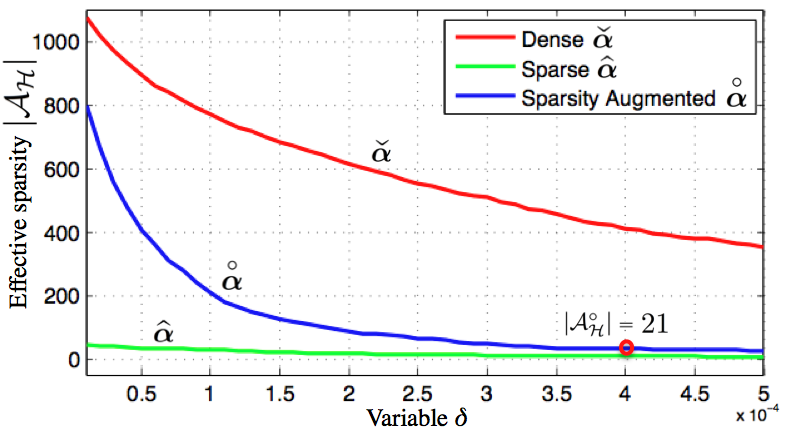} 
    \caption{Comparison of effective sparsity for face recognition, using Extended YaleB database~\cite{EYaleB}.}
    \label{fig:sparsity}
 \end{figure}

Notice that, the augmentation in Eq.~\ref{eq:alpha} also results in $|\mathcal{A^{\circ}_H}| \ll |\mathcal{A^{\smvee}_H}|$,  because the procedure reduces the relative energy in the un-amplified coefficients of $\overset{\circ}{\boldsymbol\alpha}$. 
To illustrate the difference between the effective sparsity levels of the dense and the augmented representations, we plot the effective sparsity of the representations as a function of $\delta$ in Fig.~\ref{fig:sparsity}. 
The plot is for actual face recognition task using Extended YaleB database~\cite{EYaleB}.
The curve for the augmented representation remains significantly lower than the curve for the dense representation. 
Moreover, for $\delta > 3\times10^{-4}$, $\overset{\circ}{\boldsymbol\alpha}$ is effectively almost as sparse as $\widehat{\boldsymbol\alpha}$.

Considering the definition of effective sparsity, ideally, the coefficients of $\overset{\circ}{\boldsymbol\alpha}$ indexed in $\mathcal{A_L^{\circ}}$ must be forced to zero before using the representation for classification.
However, since $\delta$ is unknown, identifying the exact $\mathcal{A_L^{\circ}}$ remains NP-hard.
To resolve this issue, we design the labeling criterion that largely remains insensitive to the coefficients indexed in $\mathcal{A_L^{\circ}}$.
That is, instead of deciding the class label of a test sample based on the fidelity of its reconstruction, we directly integrate the coefficients of $\overset{\circ}{\boldsymbol\alpha}$ for each class separately.
The largest integrated value indicates the correct class label.
Due to the dominance of large values of the coefficients of the correct class in $\overset{\circ}{\boldsymbol\alpha}$, $\mathcal{A_L^{\circ}}$ is not able to strongly influence the classification results. 
More precisely,  
our classification result 
remains as reliable as that obtained using an accurate representation with sparsity level $|\mathcal{A_H^{\circ}}|$, under   the mild worst-case condition $\sum_a - \sum_{b} > 2\sqrt{\delta}(n_b - n_a)$. Here, $\sum_a$ and $\sum_{b}$ denote the largest and the second largest integrated values of the coefficients, respectively, and $n_b$ and $n_a$ are the number of coefficients in $\overset{\circ}{\boldsymbol\alpha}$ contributing to $\sum_b$ and $\sum_{a}$ respectively, such that, each coefficient has energy less than $\delta$.
To exemplify, in Fig.~\ref{fig:sparsity}, the classification results are as accurate as possible with sparsity level $21$, unless $\sum_a - \sum_{b} \leq 0.04\times(n_b - n_a)$. 
Typically, $\sum_a - \sum_{b}\in [0.1~0.3]$, whereas $n_b \approx n_a$.
Since our labeling criterion does not need to compute reconstruction residuals for each class, we directly use the matrix ${\bf L}$ in step (3) of Algorithm~\ref{alg:2}. 
The matrix multiplication ${\bf L} \overset{\circ}{\boldsymbol\alpha}$ simultaneously integrates the coefficients for each class.
Computationally, this makes our labeling step extremely effective.

\section{Experiments}
\label{sec:Exp}
We evaluated the proposed approach using two face databases: AR database~\cite{AR} and Extended YaleB~\cite{EYaleB}, an object category database: Caltech-101~\cite{C101} and an action dataset: UCF sports actions~\cite{UCF}.
These datasets are commonly used to benchmark the approaches that use collaborative representation for  classification.
We compare the performance of our approach to SRC~\cite{SRC}, CRC-RLS~\cite{CRC}, LC-KSVD~\cite{LCKSVD}, D-KSVD~\cite{DKSVD}, FDDL~\cite{FDDL} and DL-COPAR~\cite{DLCOPAR}.  
Unless mentioned otherwise, we performed our own experiments using the same training and testing partitions for all the approaches including the proposed approach.
We carefully optimized the parameter values of the approaches using cross validation.
For the existing techniques, these values are generally the same as those reported in the original works. 
However, for some cases, we used different values to favors these approaches. 
We explicitly mention these differences. 
For the dictionary learning approaches, the dictionaries are learned using the same training data that is directly used by SRC, CRC-RLS and the proposed approach.

We used the \emph{author-provided codes} for CRC-RLS, LC-KSVD, FDDL and DL-COPAR.
For SRC, we used the SPAMS toolbox~\cite{SPAMS} to solve the $\ell_1$-norm minimization problem.  
For D-KSVD, we modified the public code of LC-KSVD~\cite{LCKSVD}.
In all the experiments, the proposed approach uses the implementation of OMP made public by Elad et al.~\cite{BOMP}.
The same implementation is used by LC-KSVD and D-KSVD.
The proposed approach uses the sparsity threshold $ k = 50$ for all the datasets.
The regularization parameter $\lambda$ is set to $0.003$ for the face databases, $1.0$ for the object database and $0.01$ for the action database.
Experiments are performed on an Intel Core i7-2600 CPU at 3.4 GHz with 8~GB RAM.



\subsection{AR Database}
\label{sec:ARD}
The AR database~\cite{AR} consists of over $4,000$ face images of $126$ subjects.
For each subject, $26$ images are taken during two different sessions with large variations in terms of facial disguise, illumination and expressions.
For our experiments, a $165\times 120$ image was projected onto a $540$-dimensional vector using a random projection matrix. 
Thus, the used samples are the Random-Face features~\cite{SRC}.   
We followed a common experimental protocol by selecting a subset of $2,600$ images of $50$ male and $50$ female subjects from the database.
For each subject, $20$ random images were chosen to create the training data and the remaining images were used for the test data.

In Table~\ref{tab:AR}, we summarize the results on the AR~database.
The reported accuracies are the means (and standard deviations) of ten experiments. 
We also report the average time taken by each approach to classify a single test sample.
For the parameter values of DL-COPAR, we followed the face recognition parameter settings in \cite{DLCOPAR}, which uses  $15$ atoms per class to represent class-specific data and $5$ atoms to represent the commonalities. 
The Local Classifier~\cite{DLCOPAR} resulted in the best accuracy for DL-COPAR.
For LC-KSVD~\cite{LCKSVD} and D-KSVD~\cite{DKSVD} we set the sparsity threshold to $50$ and the dictionary size to $1510$ atoms for improved results. 
These values are different from the original works because these were found to give the best accuracies.
For FDDL, we used the same parameter settings as~\cite{FDDL} and the Global Classifier resulted in the best performance.   
For SRC~\cite{SRC}, we set the error tolerance $\varepsilon = 0.05 $, as in the original work.
For CRC-RLS~\cite{CRC}, the regularization parameter $\lambda$ is set to $0.003$.
This value is computed using the formula provided for $\lambda$ for the face databases in \cite{CRC}. 
Our cross-validation verified that this value results in the best performance of CRC-RLS.

Table~\ref{tab:AR} shows that the best results are achieved by the proposed approach, i.e.~SA-CRC.
We have also shown the results of our approach when we use only the Regularized Least Squares (RLS) or OMP in Algorithm~\ref{alg:2}.
It is clear that using the augmented vector is better than using any of the two representation vectors alone.
Notice that, due to the efficient classification criterion, our approach is much faster than CRC-RLS even when both OMP and RLS are used.
The dictionaries used by LC-KSVD and D-KSVD are smaller in size as compared to the one used by SA-CRC, which results in slight computational advantage for these approaches. 
Nevertheless, accuracies of these approaches are much lower than SA-CRC. 

 \begin{table}[t]
  \caption{Recognition accuracies on the AR database \cite{AR} using Random-Face features. The reported average time (in milli-seconds) is for classifying a single test sample.}
  \centering
\begin{tabular}{| l | c | c |}
  \hline                       
  Method & Accuracy ($\%$) & Time  \\
  \hline \hline
    DL-COPAR \cite{DLCOPAR} & $ 93.33 \pm 1.69$ & $40.01$  \\
      LC-KSVD \cite{LCKSVD} & $ 95.20 \pm 1.22$ & $\hspace{1.5mm}1.56$  \\
       D-KSVD \cite{DKSVD} & $95.41 \pm 1.43$ &  $\hspace{1.5mm}1.54$ \\
    FDDL \cite{FDDL} & $ 96.24 \pm 1.01$ & $51.23$  \\
  SRC \cite{SRC} & $ 96.51 \pm 1.36$ & $69.91$  \\
  CRC-RLS\cite{CRC} & $ 97.65 \pm 0.67$ & $\hspace{1.5mm}4.46$ \\
  SA-CRC (only RLS) & $97.13 \pm 0.74$ & $\hspace{1.5mm}  0.07$ \\
  SA-CRC (only OMP) &$ 97.25\pm 0.43$ &$\hspace{1.5mm}2.00$ \\
  SA-CRC (proposed) & ${\bf 98.29 \pm 0.46}$ & $\hspace{1.5mm}2.13$ \\
  \hline
  \end{tabular}
\label{tab:AR}
\end{table}
 \begin{table}[t]
  \caption{Performance gain with SA-CRC in dictionary learning based multi-class classification. The average time (in milli-seconds) is for classifying a single test sample.}
  \centering
\begin{tabular}{| l | c | c |}
  \hline                       
  Method & Accuracy ($\%$) & Time \\
  \hline \hline
  K-SVD \cite{KSVD} + Lin. Classifier 		& $ 94.06\pm 1.03$ & $\hspace{1.5mm}1.56$   \\
     K-SVD $\boldsymbol\Phi$ $\rightarrow$ SA-CRC 						& ${\bf 95.65 \pm 0.66}$ & $\hspace{1.5mm}1.61$ \\
  ODL \cite{ODL} + Lin. Classifier			& $94.60\pm 0.78$ &$\hspace{1.5mm} 1.59$ \\
   ODL $\boldsymbol\Phi$ $\rightarrow$ SA-CRC							& ${\bf 95.33\pm 0.68}$ &$\hspace{1.5mm} 1.62$\\
    LC-KSVD~\cite{LCKSVD} 	& $95.31\pm 1.06$ &$\hspace{1.5mm} 1.55$\\
     LC-KSVD $\boldsymbol\Phi, {\bf L}$ $\rightarrow$ SA-CRC 	& ${\bf 96.44\pm 0.99}$ &$\hspace{1.5mm} 1.61$\\
  \hline  
 \end{tabular}
\label{tab:AR1}
\end{table}

In Table~\ref{tab:AR1}, we demonstrate the potential of SA-CRC for improving the performance of dictionary learning based multi-class classification approaches.
The results are the mean values computed over ten experiments.
We obtained the results in the first row as follows. 
First, K-SVD~\cite{KSVD} is used to learn a dictionary containing $15$ atoms per class.
The sparse codes of the training data over the dictionary are used to compute a linear classifier, following~\cite{LCKSVD}.
A test sample is classified by first sparse coding it over the dictionary and then classifying its codes using the classifier. 
In the second row, we feed the same dictionary to SA-CRC as input $\boldsymbol\Phi$ instead (the test data remained the same).
We also repeated the above procedure using the Online Dictionary Learning (ODL) approach~\cite{ODL} in place of K-SVD.
The corresponding results are also reported. 
We can see a clear gain in the classification accuracies using SA-CRC in both cases.
The table also compares the classification performance of LC-KSVD~\cite{LCKSVD} with its enhancement using SA-CRC.
For the enhancement, we replaced the classification stage of LC-KSVD with SA-CRC. That is, the dictionary and the weight matrix learned by LC-KSVD are directly used as SA-CRC's inputs $\boldsymbol\Phi$ and ${\bf L}$, respectively. 
There is a clear improvement in the classification performance of LC-KSVD after this modification. 
The performance of other discriminative dictionary learning approaches can also be improved  using SA-CRC. 
Results in Table~\ref{tab:AR} and~\ref{tab:AR1} demonstrate the potential of sparsity augmented collaborative representation for improved CR-based classification across the board.

%

\subsection{Extended YaleB}
\label{sec:EYB}
The Extended YaleB face database~\cite{EYaleB} comprises $2,414$ images of $38$ subjects.
Each subject has about $64$ samples acquired under varying illumination conditions with different expressions. 
For this database, $192\times168$ cropped images were projected onto a $504$-dimensional vector to obtain the Random-Face features.
For evaluation, we used a common experimental setting, where half of the available features of each subject were used for the training data and the remaining half were used in testing.
\begin{table}[t]
  \caption{Recognition accuracies with Random-Face features on the Extended YaleB database \cite{EYaleB}. The average time (in milli-seconds) is for classifying a single sample.}
  \centering
\begin{tabular}{| l | c | c |}
  \hline                       
  Method & Accuracy ($\%$) & Time  \\
  \hline \hline
  D-KSVD \cite{DKSVD} & $94.71 \pm0.45$ &$\hspace{1.5mm}0.41$ \\
  DL-COPAR \cite{DLCOPAR} & $94.87 \pm 0.55$ & $ 31.75$\\
  LC-KSVD \cite{LCKSVD} & $95.38\pm0.64$ &$\hspace{1.5mm}0.42$\\
  FDDL \cite{FDDL} & $96.19\pm 0.71$ & $58.19$\\
  SRC \cite{SRC} & $97.06\pm0.41$ &$68.12$\\
  CRC-RLS \cite{CRC} & $97.81\pm 0.44$ & $\hspace{1.5mm}2.41$\\
  SA-CRC & ${\bf 98.32 \pm 0.43}$ & $\hspace{1.5mm}1.23$ \\
  \hline  
\end{tabular}
\label{tab:EYaleB}
\end{table}

In Table~\ref{tab:EYaleB}, we show the results on Extended YaleB.
For D-KSVD and LC-KSVD we used 600 dictionary atoms as they gave the best accuracies. 
The remaining parameters of these algorithms were set to the original values reported in \cite{LCKSVD}.
We set the regularization parameter of CRC-RLS to $0.002$ for this database, as guided by~\cite{CRC} and dictated by cross-validation.
For the remaining approaches, the parameter values reported in Section~\ref{sec:ARD} also resulted in their best performances for this database, hence they were kept the same. 
Again, SA-CRC is able to outperform the existing techniques. 
Although SA-CRC attains only a slight advantage over CRC-RLS in terms of accuracy for this dataset, it is able to classify a test sample \emph{almost twice} as fast as CRC-RLS.


\subsection{Caltech-101}
Caltech-101 database~\cite{C101} contains $9,144$ images from $101$ object classes and one class of background images. 
The classes include diverse categories of object (e.g. trees, minarets, signs) with significant shape variation within a category.
For each class, the number of available images vary between  $31$ and $800$.
In our experiments, the used image feature descriptors were obtained by  the following procedure. 
First, the SIFT descriptors~\cite{SIFT} were extracted from $16\times 16$ patches. 
Based on these descriptors, spatial pyramid features~\cite{SPM} were extracted with $1\times 1$, $2\times 2$ and $4 \times 4$ grids. 
For extracting these features, the codebook was trained using k-means, where $k = 1024$. 
Finally, the dimension of a feature was reduced to $3,000$ using PCA.
Following a common experimental setting, we created 5 sets of train and test data with the extracted features.
These sets consisted of  $10, 15, 20, 25$ and $30$ training samples per class, whereas the remaining samples were used as the test data in each case.
We repeated our experiments ten times, every time selecting the training and testing data randomly.

Table~\ref{tab:C101} shows the mean classification accuracies for our experiments. 
We used the error tolerance of $10^{-6}$ for SRC, which gave the best results.
The regularization parameter $\lambda = 1.0$ for CRC-RLS.
The same value of the regularization parameter is used by SA-CRC to solve the RLS problem.
For FDDL, we used the parameter settings of the object categorization   experiments in \cite{FDDL}.
DL-COPAR and LC-KSVD use the same settings as in the original works for the same database.
These settings also resulted in their best performance for our data.
For D-KSVD, the settings used by \cite{LCKSVD} showed the best results.

It is clear from Table~\ref{tab:C101} that SA-CRC consistently outperforms the existing approaches. 
In Table~\ref{tab:C101Time}, we also report the classification time (for the complete test data) of the four most efficient approaches.
The time is computed when $30$ samples were used for training and the rest were used for testing.
We can see that SA-CRC is more than six times faster than CRC-RLS and its timings are comparable to those of the efficient discriminative dictionary learning approaches.
Note that, D-KSVD and LC-KSVD also required around $90$ minutes of training. 
 
\begin{table}[t]
  \caption{Classification accuracies (\%) on the Caltech-101 dataset~\cite{C101} using the spatial pyramid features. }
  \centering
\begin{tabular}{| l | c | c | c | c | c | c |}
  \hline                       
  Training samples & 10 & 15 & 20 & 25 & 30 \\
  \hline \hline
SRC \cite{SRC}  & 57.8 & 63.3 & 67.2 & 69.2 & 71.8 \\
CRC-RLS \cite{CRC}  & 59.4 & 64.8 & 68.0 & 69.3 & 71.8 \\
DL-COPAR \cite{DLCOPAR} & 58.4 & 65.1 & 69.3 & 71.1 & 72.5\\
FDDL \cite{FDDL}  & 59.7 & 66.6 & 69.1 & 71.3 & 72.9\\
D-KSVD \cite{DKSVD} & 60.7 & 66.3 & 69.6 & 71.0 & 73.1 \\
LC-KSVD \cite{LCKSVD}  & 62.9 & 67.3 & 70.3 & 72.6 & 73.4 \\
SA-CRC  & {\bf 63.2} & {\bf 68.2} & {\bf 71.9} & {\bf 73.6} & {\bf 76.1} \\
  \hline  
\end{tabular}
\label{tab:C101}
\end{table}

  \begin{table}[t]
  \caption{Computation time (in seconds) for classification on  Caltech-101 dataset~\cite{C101}. }
  \centering
\begin{tabular}{| l | c || l | c |}
\hline
Method & Time & Method & Time\\
  \hline\hline
 D-KSVD \cite{DKSVD} 		& 19.80  			& SA-CRC			 	& 21.43\\
 LC-KSVD \cite{LCKSVD} 	& 19.91			& CRC-RLS \cite{CRC}	& 130.41\\
 \hline
  \end{tabular}
  \label{tab:C101Time}
  \end{table}
  
 \subsection{UCF Sports Actions}
 The UCF Sports Action dataset~\cite{UCF} contains video sequences collected from different broadcast sports channels.
 The videos are from $10$  categories of sports actions (e.g. diving, lifting, running).  
For this dataset, we directly used the action bank features made public by Sadanand and Corso~\cite{actBank}.
A common evaluation protocol was followed in our experiments, where a  fivefold cross validation was performed using four folds for training and the remaining one for testing.
The results in Table~\ref{tab:UCF} are the average accuracies of the five experiments.
The reported accuracies of  Sadanand~\cite{actBank}, DL-COPAR and FDDL are taken directly from~\cite{LDL}, where the same experimental protocol has been followed. 
Our parameter optimization for FDDL and DL-COPAR could not achieve these accuracies.
For the remaining  approaches the results are reported on the same folds using the optimized parameter values.
For SRC, the error tolerance was set to $10^{-6}$ and $50$ dictionary atoms were used for LC-KSVD and D-KSVD.
The same number of atoms were used by Jiang et al.~\cite{LCKSVD}.
We used $\lambda = 0.01$ for both CRC-RLS and SA-CRC, which resulted in their best performance.
For all the five experiments combined, the classification time was $0.04$ seconds for SA-CRC and $0.31$ seconds for CRC-RLS.

\begin{table}[t]
  \caption{Classification accuracies (\%) on UCF Sports Action dataset~\cite{UCF} using the action bank features.}
  \centering
\begin{tabular}{| l | c || l | c |}
\hline
Method & Acc. & Method & Acc.\\
  \hline\hline
Sadanand \cite{actBank} 		& $90.7$ 		& FDDL \cite{FDDL} 			& $93.6$ \\
DL-COPAR \cite{DLCOPAR} 	& $90.7$		& LC-KSVD \cite{LCKSVD}	& $94.2$  \\
D-KSVD \cite{DKSVD} 			& $93.4$  		& CRC-RLS \cite{CRC} 		& $94.4$  \\
SRC \cite{SRC} 					& $93.5$		& SA-CRC 					& ${\bf 95.7}$\\
 \hline
  \end{tabular}
  \label{tab:UCF}
  \end{table} 
\section{Discussion}
\label{sec:Disc}
\begin{figure*}[t] 
   \centering
   \includegraphics[width=7in]{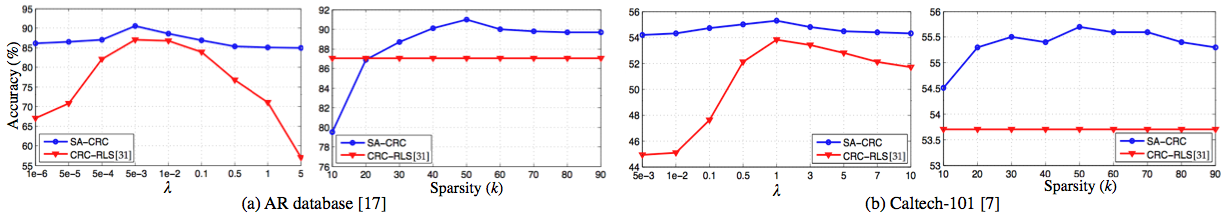} 
   \caption{Accuracy as a function of parameters: One parameter is fixed for SA-CRC while the other is varied. Both CRC-RLS~\cite{CRC} and SA-CRC always use the same values of $\lambda$. (a) $k$ is fixed to $50$ in the first plot (from left) and $\lambda$ is fixed to $0.003$ in the second. (b) $k$ is fixed to $50$ and $\lambda$ to $1$, respectively. }
   \label{fig:para}
\end{figure*}
Our approach requires a regularization parameter $\lambda$ and sparsity threshold $k$ as the input parameters for a  given pair of  $\boldsymbol\Phi$ and its label matrix~${\bf L}$. 
In our experiments, we optimized the values of these parameters by cross-vlaidation using the following systematic procedure.
First, $\lambda$ was optimized by executing Algorithm~\ref{alg:2} without step~1(b) and considering $\widehat{\boldsymbol\alpha}$ to be a zero vector in Eq.~\ref{eq:alpha}.  
Then $k$ was optimized by fixing $\lambda$ to the optimized value and executing the complete Algorithm. 
The parameters were further fine-tuned to nearby values when  doing so yielded better performance.

To show the behavior of SA-CRC for different parameter values,    in Fig.~\ref{fig:para}, we plot the classification accuracy of SA-CRC as a function of $\lambda$ and $k$ by fixing one parameter and varying the other. 
We also include results of CRC-RLS~\cite{CRC} for comparison. 
Plots in Fig.~\ref{fig:para}a, are for AR database~\cite{AR} where we followed the experimental protocol of \cite{CRC}.  
In the first plot (from left), we fixed $k$ to $50$ and varied $\lambda$. 
Clearly, SA-CRC consistently outperforms CRC-RLS and the results are less sensitive to the values of $\lambda$ once $k$ is fixed to an optimized value. 
In the second plot, we used $\lambda = 0.003$ for both SA-CRC and CRC-RLS and varied $k$ for SA-CRC. 
Again, for $k > 20$, SA-CRC consistently outperforms CRC-RLS.
Qualitatively speaking, Fig.~\ref{fig:para}a shows a typical relationship between the performance of CRC-RLS and SA-CRC that was observed in our experiments on face databases. 

In Fig.~\ref{fig:para}b, we repeated the same experiment for the object dataset, Caltech-101~\cite{C101}.
To fix the parameter values, we used $k = 50$ and $\lambda = 1$.   
For this experiment, we used five samples per class for training and the rest for testing. 
Again, the obtained results consistently favor SA-CRC.
Qualitatively similar behavior  was observed for all the train/test partitions used in Table~\ref{tab:C101}. 


\section{Conclusion}
\label{sec:Conc}

In contrast to a popular existing notion, we showed that sparsity of a Collaborative Representation (CR) plays an explicit role in accurate CR-based classification, hence it should not be completely ignored for computational gains.
Inspired by this result, we propose a Sparsity Augmented Collaborative Representation based Classification scheme (SA-CRC) that augments a dense collaborative representation with an efficiently computed sparse representation. 
The resulting representation is classified using a efficient method.
Extensive experiments for face, action and  object classification establish the effectiveness of SA-CRC in terms of accuracy as well as computational efficiency.

\ifCLASSOPTIONcompsoc
  \section*{Acknowledgments}
\else
  \section*{Acknowledgment}
\fi

This work was supported by ARC Grants DP110102399 and DP1096801.

\ifCLASSOPTIONcaptionsoff
  \newpage
\fi



\balance
{\small
\bibliographystyle{ieee}
\bibliography{MyTex}
}

%




%




\end{document}